# THE DYNAMICS OF THE STOMATOGNATHIC SYSTEM FROM 4D MULTIMODAL DATA

**Agnieszka A. Tomaka, Leszek Luchowski, Dariusz Pojda, Michał Tarnawski, Krzysztof Domino**

Institute of Theoretical and Applied Informatics, Polish Academy of Sciences

**ABSTRACT**

Purpose

The purpose of this chapter is to discuss methods of acquisition, visualization and analysis of the dynamics of a complex biomedical system, illustrated by the human stomatognathic system.

Material and Methods

The stomatognathic system consists of the teeth and the skull bones with the maxilla and the mandible. Its dynamics can be described by the change of mutual position of the lower/mandibular part versus the upper/maxillary one due to the physiological motion of opening, chewing and swallowing. In order to analyse the dynamics of the stomatognathic system its morphology and motion has to be digitized, which is done using static and dynamic multimodal imagery like CBCT and 3D scans data and temporal measurements of motion. The integration of multimodal data incorporates different direct and indirect methods of registration – aligning of all the data in the same coordinate system.

Results

The integrated sets of data form 4D multimodal data which can be further visualized, modeled, and subjected to multivariate time series analysis. Example results are shown.

Conclusions

Although there is no direct method of imaging the TMJ motion, the integration of multimodal data forms an adequate tool. As medical imaging becomes ever more diverse and ever more accessible, organizing the imagery and measurements into unified, comprehensive records can deliver to the doctor the most information in the most accessible form, creating a new quality in data simulation, analysis and interpretation.

## 1. INTRODUCTION

The present chapter presents the methods of acquisition, visualization and analysis of the dynamics of a biological system from 2, 3, and 4D multimodal data. Multimodal in this context means using various types of imaging devices, based on different physical and geometric principles, whose powers and limitations complement each other.

Medical imaging is a good example of multimodal imagery. The reality of a living organism is four-dimensional, defined in the three dimensions of our space and the fourth dimension of time. Imaging devices examine the spatial distribution [20] of certain physical parameters either in the full volume of the object (such as X-ray attenuation coefficient) or of its surface (like the Lambertian reflectance seen by photo cameras). Some imaging devices are able to perceive a two-dimensional projection of the 3D distribution; such is the case of traditional X-ray and ordinary photography. Some others perceive the distribution in three dimensions, as do various radiological computed tomographs and optical 3D scanners. When 3D imaging ability is repeated in a time series, the resulting data is referred to as a 4D image (3D + time). Technically, a non-stereoscopic video sequence might be called a 3D image, but we shall not use 3D in this sense, to avoid confusion.

In this chapter we concentrate on data acquired from the human stomatognathic system (SS), as this system is of considerable medical importance as well as being complex and difficult to examine.
The human SS [10] includes the maxilla (upper jaw) with adjoining cranial bones and the mandible (lower jaw). These two bone structures are connected at the two temporomandibular joints (TMJs) on either side of the head. The position of the jaws at rest determines the occlusion, i.e. the way the teeth come together. The dynamics of the SS can be described by the change of mutual position of the lower/mandibular part with respect to the upper/maxillary one due to the physiological motion of opening, chewing and swallowing. From a medical point of view the relation between upper and lower parts is particularly important in the areas of tooth contact between the two arches. The same relation manifests itself also as the position of the condyles in the articular fossa in the TMJ [14].

Modern imaging techniques offer the possibility to obtain the shape of the bones and teeth of the stomatognathic system. It is also possible to measure the mandible movement [26], and to take a 3D photograph of the patient's face. All of these images, image sequences, and geometric data are represented in various formats, in different coordinate systems, and at different points in time. The aim of the present chapter is to show the methods of integrating such an assortment of data into a 4D multimodal record representing the changes of the maxilla-mandibular relations occurring during physiological motion.

Medically, the geometry and dynamics of the SS is relevant for a number of reasons:
- A large class of disorders are associated with situations where maximal intercuspation of teeth does not occur in the same position of the jaws as the centric relation at the two TMJs [15]. Maximal intercuspation is an occlusion where the cusps of teeth interlock most closely, while centric relation means the optimal position of the heads of the mandibular condyles in the articular fossa. Imaging and quantifying these relations is an essential diagnostic instrument.
- The design of intraoral appliances [1][5] to control and correct occlusion also requires precise 3D measurements both in areas of tooth contact and in the TMJ, in the static position as well as during physiological motion
- The same is true in the treatment of bruxism (involuntary clenching and/or gnashing of teeth, often, but not always, in sleep), which can be alleviated by the use of occlusal splints [2] to redistribute pressure between the upper and lower teeth [13]

## 2. MATERIAL

Analysing the occlusion in the areas of dental contact and at the TMJ requires the bone tissue of the SS to be imaged in the habitual resting position (static occlusion) and in various phases of its motion.

### 2.1. Static morphometry

The orthodontic patient record includes different kinds of 2D/3D images acquired using visible light or X-ray based imaging devices [12]. While visible light imaging acquires only the outer patient surfaces, X-rays passing through the inner tissues are capable of creating images of invisible layers.

#### 2.1.1. CBCT

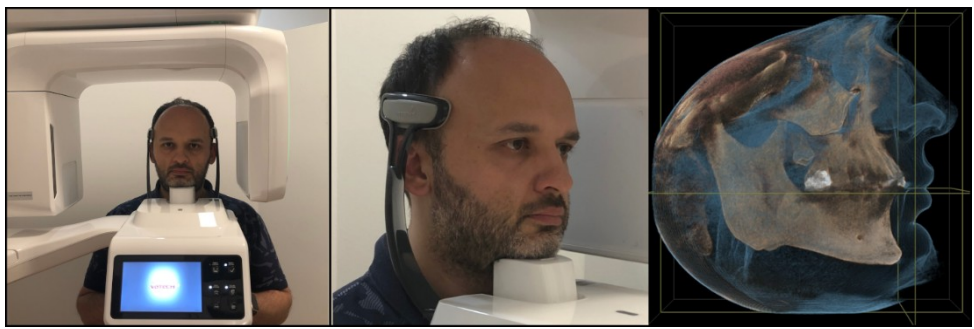

Fig. 1. Positioning a patient for CBCT examination and the obtained results

The 3D shape of the skull and mandibular bones and teeth has to be acquired using X-ray based means such as Computerized Tomography or Cone Beam Computed Tomography. CBCT (Fig.1) as a relatively new type of X-ray imaging, creating 3D images at little radiation-exposure cost to the patient, becomes a standard in orthodontic diagnosis and planning of the therapeutic interventions.

The idea of X-ray examinations relies on measuring the radiation passing through the subject being imaged. As the attenuation of X rays depends on the chemical composition of the tissues, the differences between the tissues cause differences in measured radiation, which are the basis for tissue discrimination in the resulting image. Therefore, all X-ray imaging modalities excel at imaging bone and highly mineralized components of teeth, while performing less well with details of soft tissues. As radiation is attenuated by all the tissues though which it passes on its way, the resulting image is a kind of superposition of projections of each layer. This property of X-ray projection is the basis for reconstruction of the CT image of a slice by computing the local attenuation coefficients from the set of summative projections obtained by sending X rays in different directions through the same target area. From the set of sums the space distribution of the attenuation coefficient is then reconstructed [4] [6]. Repeating the examination for different slices, a volumetric image is obtained, composed of voxels with determined attenuation coefficient.

The voxels obtained from CBCT examinations can be used to generate simulations of pantomograms and cephalograms, to create images of thin layers, and finally to reconstruct the 3D surfaces of tissues with a similar x-ray absorption coefficient (Fig. 2). The 3D surface reconstruction finds the points of the same intensity and builds the mesh which joins them.

Such 3D isosurface reconstructions obtained for different absorption coefficients yield patient specific models of teeth, bones and soft tissues. The reconstructions are obtained in the same CBCT-connected coordinate system and therefore their mutual relations reflect the relation between the real tissues. Application of the CBCT has some drawbacks connected with a limited Region of Interest (ROI) due to the need to limit radiation exposure, and problems with segmentation in the areas of contact between separate but touching parts.

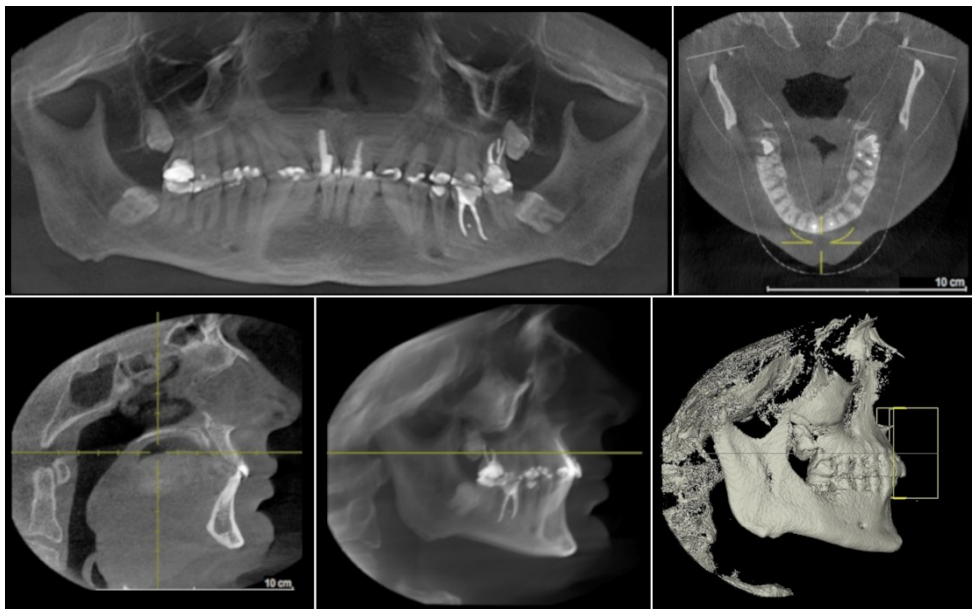

Fig. 2. Simulation of the pantomogram, axial and saggital crossections, cephalograms and 3D surface reconstruction in Galileos Viewer

**2.1.2 Visual-light 3D scanning**

Additional 3D surface scanning can be used to overcome the problems related to limited ROI and segmentation difficulties. 3D scanning can be applied to the outer surfaces of the patient's face and to the upper and lower tooth arches or to dental models. In each application either dedicated visual-light 3D scanners or a general-purpose one can be used.

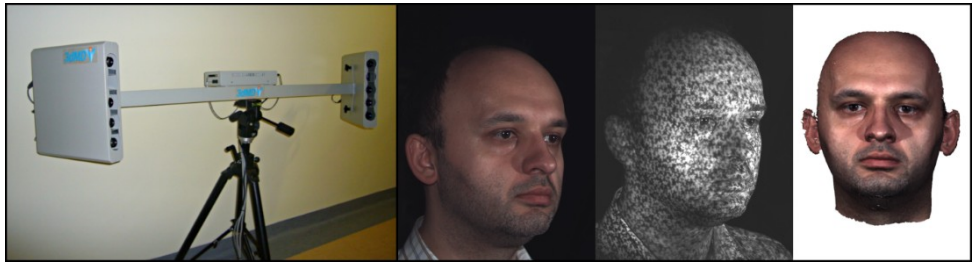

Fig. 3. 3D Scanning of a patient's face using the 3dMD face scanner, exemplary images of the cameras and the obtained result

Visual-light scanners [18][21] use visual (or infrared) light to acquire the outer shape of objects. At the scale of the human head and SS, most scanning devices use the principle of triangulation, projecting a light pattern at the object from one point and observing it by a camera situated at another point (Fig. 3). The light source, the pupil of the camera, and any identifiable part of the pattern on the surface of the object form a triangle with a known base and two angles, which allows the illuminated point to be described by 3D coordinates. Patterns of light vary between scanner types, from a single dot or a narrow stripe of monochromatic laser light (Fig 4) to a (quasi-) random array of dots to a striped motif with a sine-wave profile of greylevels. Coloured patterns have also been used. The result of scanning is a large set (tens of thousands to millions) of 3D points described by 3D coordinates, representing the surface being examined. Frequently, the points are organized into a triangular or other polygonal mesh, either by the scanner itself or by the early stages of processing software. Most scanners also register the colour of the surface at each point, which allows the 3D shape to be rendered in realistic color, known as texture. Visible-light (or infrared) scanning is noninvasive, as it does not use harmful radiation and the power of the light is within safe limits for the patient's eyes.

Visible-light scanners can be applied in three contexts in orthodontics: to scan the external shape of the face or head [23], to scan anatomic structures inside the oral cavity (intraoral scanners are built for this purpose), and to scan inanimate objects such as dental impressions, casts, or appliances [21].

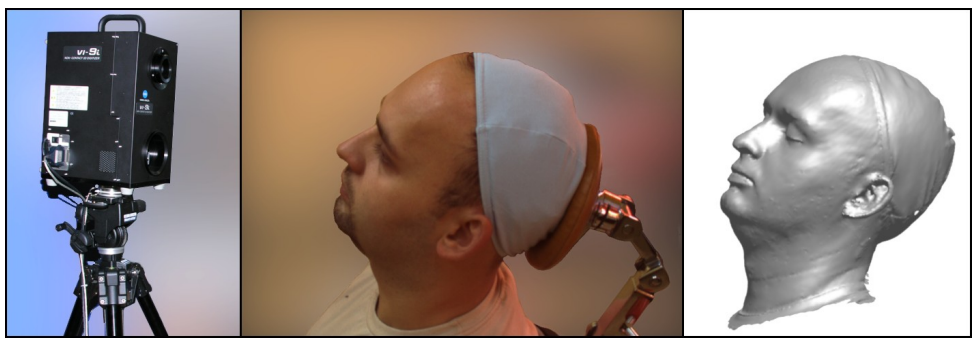

Fig. 4. 3D Scanning of a patient's face using a Konica Minolta 3D scanner. Head stabilization and the obtained result

3D scanning of a patient's face yields best results with scanners acquiring 3D images of surfaces simultaneously from different viewpoints (Fig. 3), obtaining a full image of the face/head at the same time. Merging several images taken with a scanner (Fig. 4) that was moved between different viewpoints is possible, especially for highly motivated subjects who can be expected to remain relatively motionless, but it is still prone to error due to the changes of the position of patient's soft tissues due to breathing movements.

Teeth can be scanned either directly in the patient's mouth using an intraoral scanner (Fig. 6) [17] [27], or indirectly using dental impressions (Fig 5.) or casts.

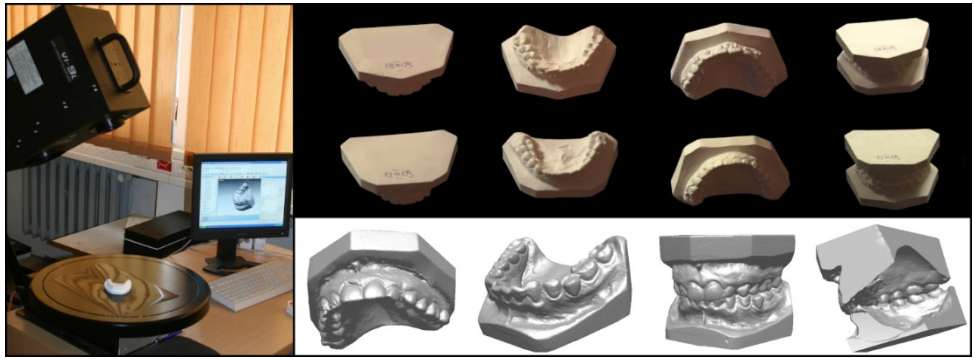

Fig. 5. 3D Scanning of a patient's dental casts using a Konica Minolta 3D scanner. Scanning positions and the obtained results of upper, lower parts, static occlusion and its cross-section

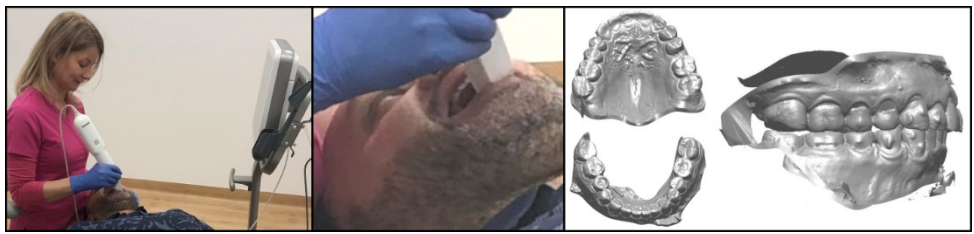

Fig. 6. 3D Scanning of a patient's dental arches using an intraoral scanner. Obtained results of upper, lower parts, and the static occlusion

### 2.2. Motion acquisition

The relation between the maxillary and mandibular part of the SS changes during normal motion of the mandible in the TMJ. The mandibular bone and teeth undergo rigid-body motion, while the soft tissues are deformed due to their flexibility. As CBCT examination cannot be repeated in each phase of mandible movement - due to the radiation exposure - other methods of tracking mandible movement have to be applied.

#### 2.2.1. Dynamic 3D scanning

The idea of using 3D scanners to track movement relies on tracking bows rigidly attached to the upper and lower bones or teeth of the stomatognathic system [25]. The face is scanned with both bows attached. The patient's head is free to move (Fig. 7), but its image is stabilized in post-processing by bringing the representations of the facial bow in all the images of the sequence to one fixed position. Then, the positions of the lower bow indicate the movement of the mandible relative to the maxilla. The presence of facial surfaces allows the data to be brought into register with the 3D surface reconstructions from CBCT. The drawback of the method is, however, a relatively low temporal resolution. Taking 7 images per second with 3dMD scanner is not enough to track the motion.

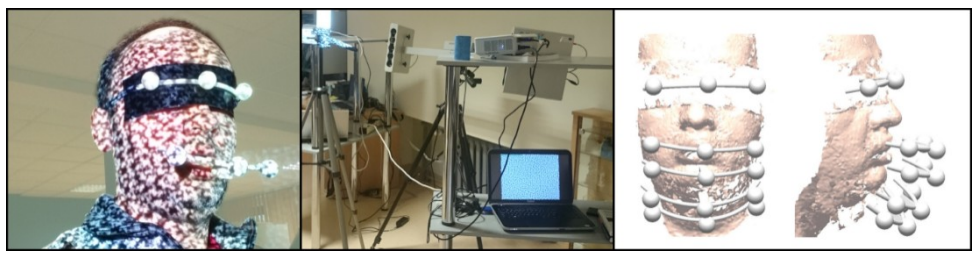

Fig. 7. 3D dynamic scanning of a patient's face with facial bows. The results after the upper bow alignment. See also [25]

### 2.2.2 Zebris axiograph movement data

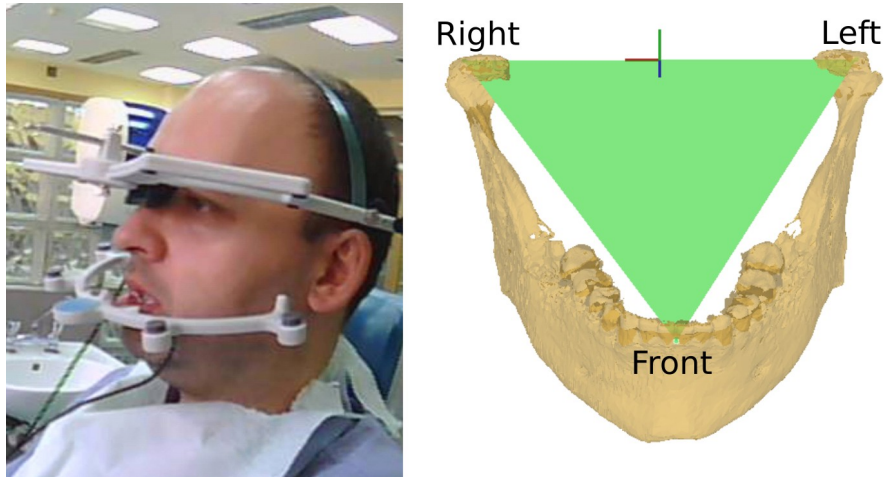

Fig. 8. Mandible motion acquisition using Zebris axiograph (left). The Bonwille triangle (right). See also [22]

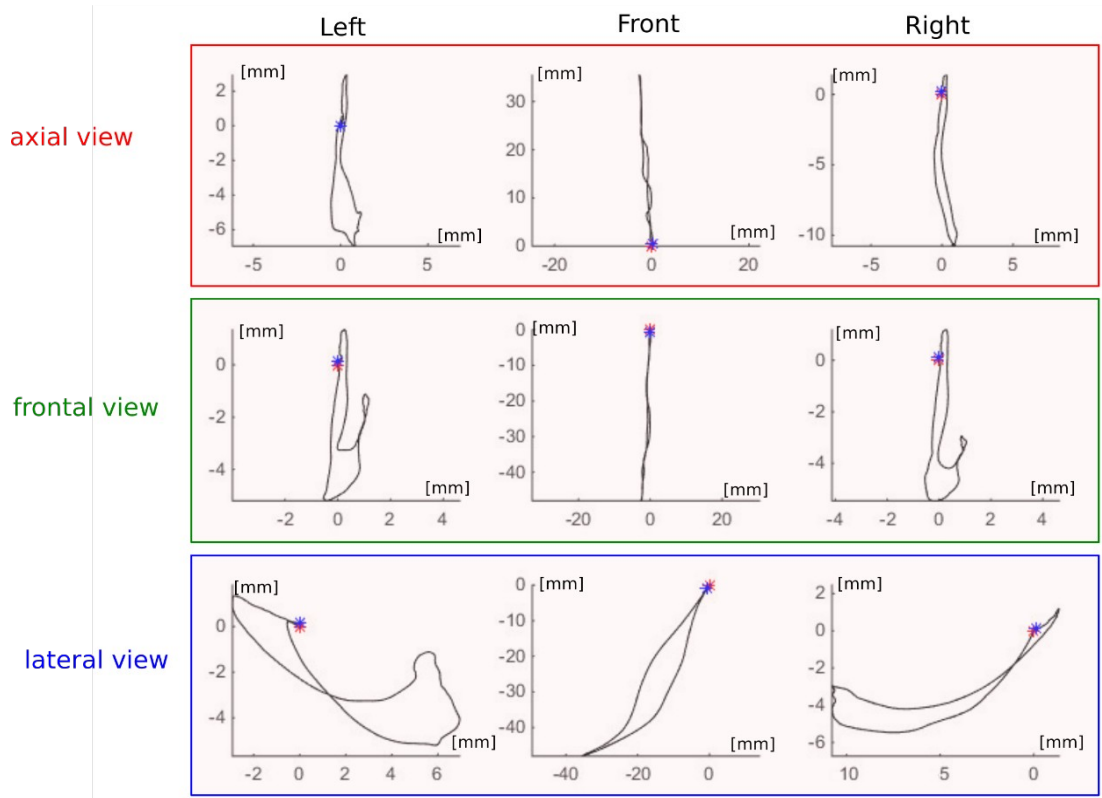

Fig. 9. Movements of the vertices of Bonwille triangle (see also [22])

The Zebris axiograph (Fig. 8) tracks the movements of the mandible relative to the skull using sets of ultrasound transmitters and receivers on two bows (frames), one of which is attached to the mandibular teeth, and the other to the patient's head. The position of the mandible is sampled 75 times per second, and reported by transmitting the position of two virtual triangles, each of which is coupled to one of the bows. The two triangles are not guaranteed to exactly correspond to any anatomical structures, just to undergo the same motion, so the motion of one triangle with respect to the other is a good representation of the motion (rotation and translation) of the mandible relative to the maxilla and the rest of the head.

## 3. METHODS

Although the motion acquisition both from the 3D dynamic scanner and the Zebris axiograph yields results which can be analysed in order to assess the symmetry and smoothness of the motion, they alone don't give the information on the conditions of the TMJ and tooth contacts.

This information for each element of the sequence only becomes available when the multimodal data can be integrated. Shown below are the various methods leading to the integration of the different kinds and representations of multimodal data.

As the data obtained from the different kinds of static and dynamic examination of the SS is expressed in device-specific coordinate systems, reducing them to a common coordinate system is prerequisite for further analysis. This process, known as registration, depends on the representation of data. The following preliminary conditions can be assumed: all image data come from the same patient, and are acquired more or less at the same time, therefore no deformations are taken into account. The images only need to be moved by rigid-body transforms, the parameters of which (rotation and translation) have to be determined.

As the various imaging devices emphasize different types of tissue, and some of them only scan the surface while others render a full 3D array of voxels, relating them to each other is not trivial. To find the transformations one needs some additional information like: geometry of imaging systems and viewpoints localization in a patient-centered coordinate system, movement of imaging device, positions of additional markers or overlapping (common) parts in the images which are to be registered.

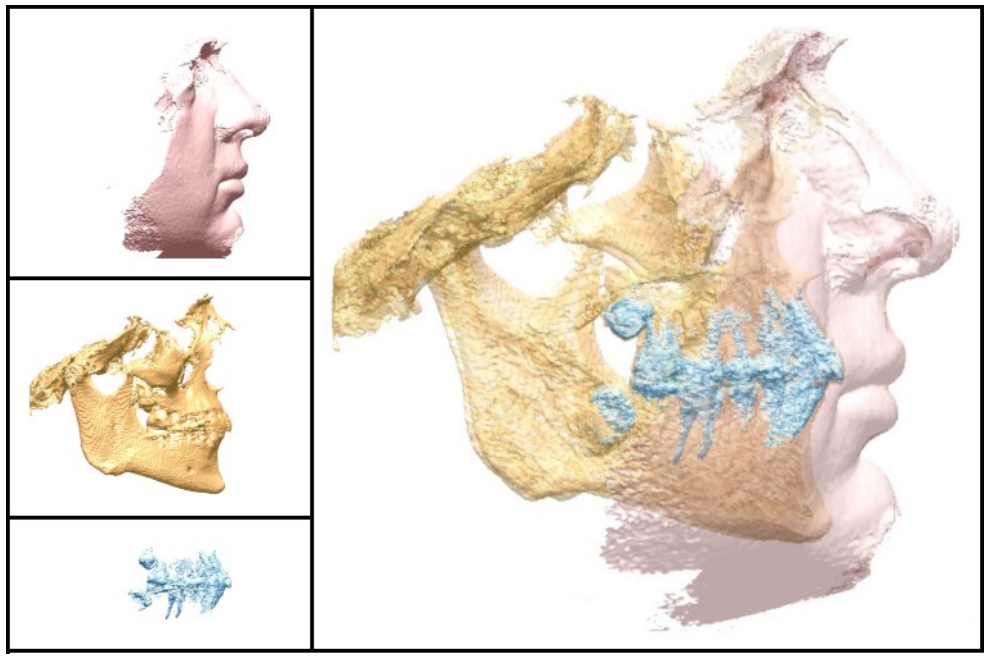

Fig. 10. Superimposition of the 3D surface reconstructions of CBCT data obtained for different attenuation coefficients (see also [22])

A simple solution is to manually indicate corresponding points on surfaces which are seen in two or more images; this approach, however, is time-consuming, labor-intensive and subject to human error.

### 3.1. The ICP algorithm – Iterative Closest Point

A general scheme for registration is as follows [11]: defining the image representation, defining a mathematical criterion of fit as an energy function, performing optimization to determine the transformation parameters, transforming the images using parameters for which the best fit is achieved according to the criterion adopted.

Determining transformation for overlapping images is connected with establishing homology (correspondence), i.e. localizing matching points in the images. If correspondences are known a priori, for example given by landmarks, the transformation that minimizes distances between corresponding points is can be calculated using a least squared method also known as the Procrustes method [9]. The rigid body transformation composed of rotation, translation is sought. Therefore, the solution can be found by minimizing the following expression, which is done by the least square method.

$$\min_{R,\gamma} D(P_1, P_2) = \|P_2 - R P_1 - \gamma 1^T\|_2$$

where: $P_1, P_2 \in R^{(3 \times k)}$ are matrices of landmarks coordinates, $R$ is a rotation matrix, $\gamma = [\gamma_x, \gamma_y, \gamma_z]$ is a translation vector, $1^T = \underbrace{[1 \ \cdots \ 1]}_{k}$ k-length vector of ones, and k is the number of points.

In the absence of information on points correspondence the iterative closest point (ICP) algorithm can be applied [3]. It assumes temporary correspondence defined for each point in the first image representation as the closest point in the second image representation. This temporary correspondence is used to determine parameters of the transformation. After the transformation, the procedure of establishing the temporary correspondence and determining transformation parameters is repeated iteratively until the sum of distances of corresponding points reaches a minimum.

The result of the registration depends on the area selected for matching and on initial mutual position of both images. For each application this area should be initially selected so as to avoid the areas with a high probability of deformation due to change of the position of the patient.

### 3.2. Indirect methods – facial bow

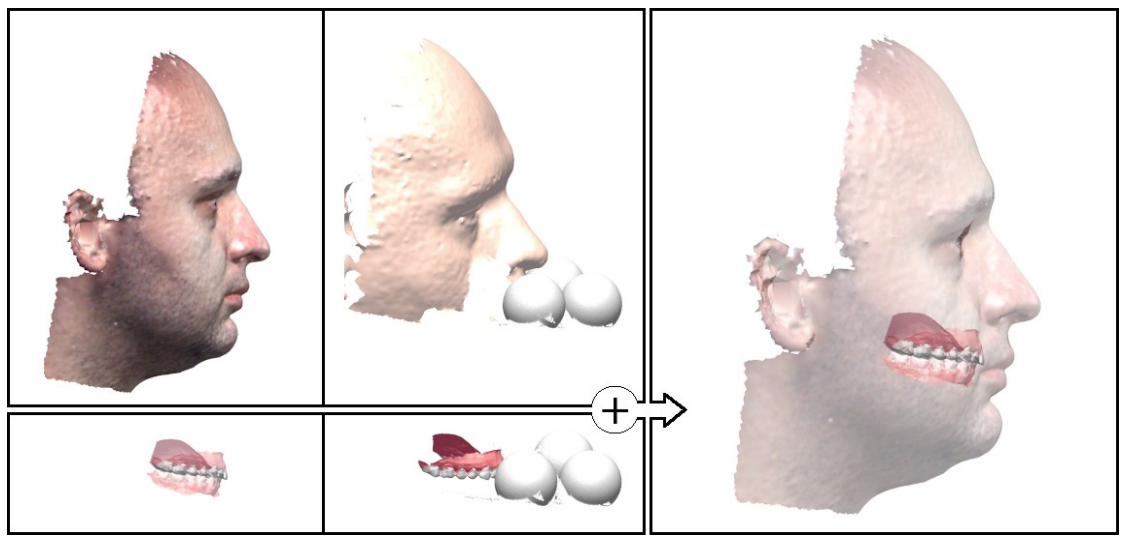

Fig. 11. Registration of disjoint 3D surfaces using an additional reference object (see also [24])

To bring into register the images of two structures that do not have common points, additional objects are required that will introduce common features to both otherwise unrelated images. Such is the case of model embedding [19][24] i.e. inserting a virtual 3D model of teeth into a 3D-scanned surface of the patient's face. These two 3D images represent the surfaces of - respectively - the patient's face and his dental models. To add common features to these two otherwise disjoint models, a facial bow is attached to an occlusal bit (piece of dental wax held, and shaped, by the patient's teeth). The bow is an easy to scan object printed by means of a 3D printer. The face is scanned twice: once with the bit held between the teeth and the bow protruding outside, and again without them. Similarly, the dental models are scanned assembled together with the bow held by the bit between them, and again separately (each model on its own). The final result only includes the images without the bow, but correctly positioned through its intermediary.

For the registration of vision-based and CBCT reconstructions the following common parts were assumed: surfaces of teeth, upper parts of face and bow surfaces. For each matching the mathematical transformation describing rigid body motion is sought. The registration can be performed pairwise, iteratively finding particular transformations, or simultaneously registering of all common parts, which finds all transformations at the same time, distributing registration error equally on each matching.

### 3.3. Registration of movements

The registration of the static maxillo-mandibular models acquired in CBCT, with the initial position of mandibular arch from the movement sequence, obtained from dynamic 3D scanner, can be performed by ICP algorithm using the upper face surfaces, which are common for both examinations.

A more complicated task is connected with the registration of the static maxillo- mandibular models with movement data acquired from the Zebris axiograph. The difficulty is connected with the localization of the triangles defined in Zebris coordinate system on the 3D surface reconstructions from CBCT. Manual localization is subject to error, although relatively small but leading to significant errors in the simulation of motion in the coordinate system bound to the static imaging data (causing not physically possible intersection of the meshes of the maxillary and mandibular parts).

A different registration technique is therefore needed. Instead of registering the sets of points as described in the previous paragraphs the registration is based on the axis representation of a rigid body transformation under the assumption of the existence of the three identical motions acquired in both coordinate systems in dynamic 3D scanner and in Zebris axiograph.

A general rigid-body transformation can be uniquely represented as a rotation around an axis and a translation along the same axis. By aligning a set of three axes, no two of them parallel, it is possible to determine the transformation aligning coordinate systems. As corresponding points are better suited to the task of coordinate system registration, an obvious choice is to use the intersections between each pair of the three lines, if any. Line pairs which do not intersect can also be used to define points; namely, on either line in a pair, the point closest to the other line can be found. These points of intersection or closest proximity between the lines are the only known point correspondence. Substituting straight line equations to the formula of rigid body transformation yields the sought rotation matrix $R$, and translation vector $\gamma$.

$$\begin{bmatrix} X_{f_1} & X_{f_2} & X_{f_3} \\ Y_{f_1} & Y_{f_2} & Y_{f_3} \\ Z_{f_1} & Z_{f_2} & Z_{f_3} \end{bmatrix} + \begin{bmatrix} a_{f_1} & a_{f_2} & a_{f_3} \\ b_{f_1} & b_{f_2} & b_{f_3} \\ c_{f_1} & c_{f_2} & c_{f_3} \end{bmatrix} t = R \left( \begin{bmatrix} X_{g_1} & X_{g_2} & X_{g_3} \\ Y_{g_1} & Y_{g_2} & Y_{g_3} \\ Z_{g_1} & Z_{g_2} & Z_{g_3} \end{bmatrix} + \begin{bmatrix} a_{g_1} & a_{g_2} & a_{g_3} \\ b_{g_1} & b_{g_2} & b_{g_3} \\ c_{g_1} & c_{g_2} & c_{g_3} \end{bmatrix} t \right) + \begin{bmatrix} \gamma_x & \gamma_x & \gamma_x \\ \gamma_y & \gamma_y & \gamma_y \\ \gamma_z & \gamma_z & \gamma_z \end{bmatrix}$$

Therefore

$$R = \begin{bmatrix} a_{f_1} & a_{f_2} & a_{f_3} \\ b_{f_1} & b_{f_2} & b_{f_3} \\ c_{f_1} & c_{f_2} & c_{f_3} \end{bmatrix} \begin{bmatrix} a_{g_1} & a_{g_2} & a_{g_3} \\ b_{g_1} & b_{g_2} & b_{g_3} \\ c_{g_1} & c_{g_2} & c_{g_3} \end{bmatrix}^{-1}$$

and

$$\gamma = \begin{bmatrix} \gamma_x \\ \gamma_y \\ \gamma_z \end{bmatrix} = \begin{bmatrix} X_{f_1} \\ Y_{f_1} \\ Z_{f_1} \end{bmatrix} - R \begin{bmatrix} X_{g_1} \\ Y_{g_1} \\ Z_{g_1} \end{bmatrix}$$

where $(X_f, Y_f, Z_f)$ and $(X_g, Y_g, Z_g)$ with subscripts are corresponding points of the least distance between axes (in a special case the one common point where all three axes cross.). $(a_f, b_f, c_f)$, $(a_g, b_g, c_g)$ with subscripts are the vectors of directional coefficients of the corresponding axes, determined from three pairs of corresponding rigid body transformations.

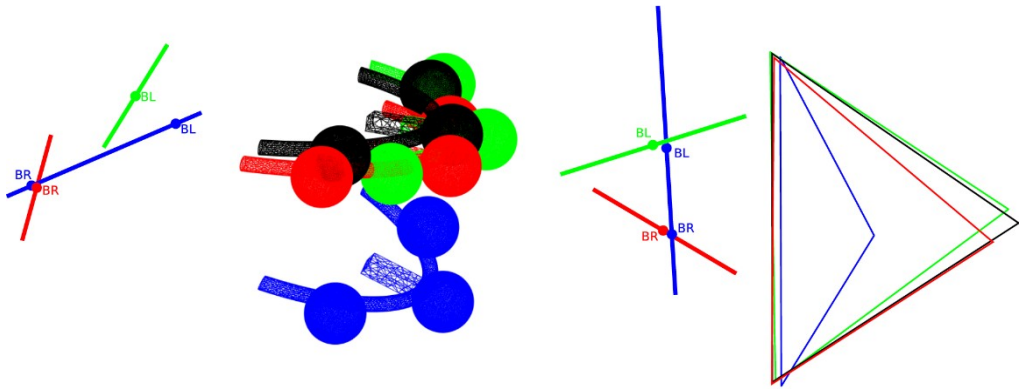

Fig. 12. Axes-based registration of movements

Fig. 12 shows the idea of registration, where the coloured balls represent the facial bow in various position of the mandible, and the straight lines are the rotation-and-translation axes transforming the reference position (black bow) to each of the other positions. The points of closest proximity are marked on the lines. Positions of the Bonville triangle and corresponding axes are also shown.

## 4. RESULTS

The overall idea of the measurement relays on the registration of the static maxillo-mandibular models acquired in CBCT, with the initial position of mandibular arch from the movement sequence. Next the relation change between arches in the sequence is measured and yielded transformation of mandibular arch is applied upon the mandible model. Obtained integrated data set of the motion sequence is the result from which the further steps can be done. The simulations of the movements can now be visualised in each phase of the motion and as the animation. The visualization can be presented as the moving surfaces, cross-sections in any view.

A virtual X-ray projection [21] is created by defining the parameters of the imaging system the position of the detector, the relation of the radiation source to the object being examined, stating the type of the source (point or extended), defining the energy to be used. For a voxel representation, creating the projection amounts to computing the sum of densities of the voxel intersected by the ray reaching a given 2D image pixel. Unfortunately, a voxel representation is only available for the position in which the CBCT was taken. In simulations using 3D surface scans the X-ray attenuation coefficient in voxels inside the surface is not given; in such cases, a standard value of the coefficient for each radiation energy has to be assumed throughout the volume in question. For each ray passing through the volume, the length of its passage is computed and multiplied by the attenuation coefficient. The simulation can be repeated for various radiation energies, influencing the tissue discriminating power of the virtual X-ray image.

The occlusion is determined as the contact points between two disjoint bodies. In practice, because of the measurement precision of the scans and reconstructions, search for exact contact points – with distance zero – is meaningless. A different approach has to be used, identifying points where the distance between the bodies is less than a certain threshold, i.e. 0.5 mm [22]. Mesh topology must also be taken into account, to ensure that a vector connecting closest points on the two bodies does not pass through the interior of either of them.

Performing virtual X-ray projections of the stomatognathic system can yield the sequence of planar virtual X-rays showing the movement of the condylion in the TMJ. The analysis of the tooth areas of contact between arches gives the insight on the occlusion which occurs in the particular phases of the motion, which can be helpful to identify non proper contacts, asymmetry and other disorders in the stomatognathic system (Fig. 13).

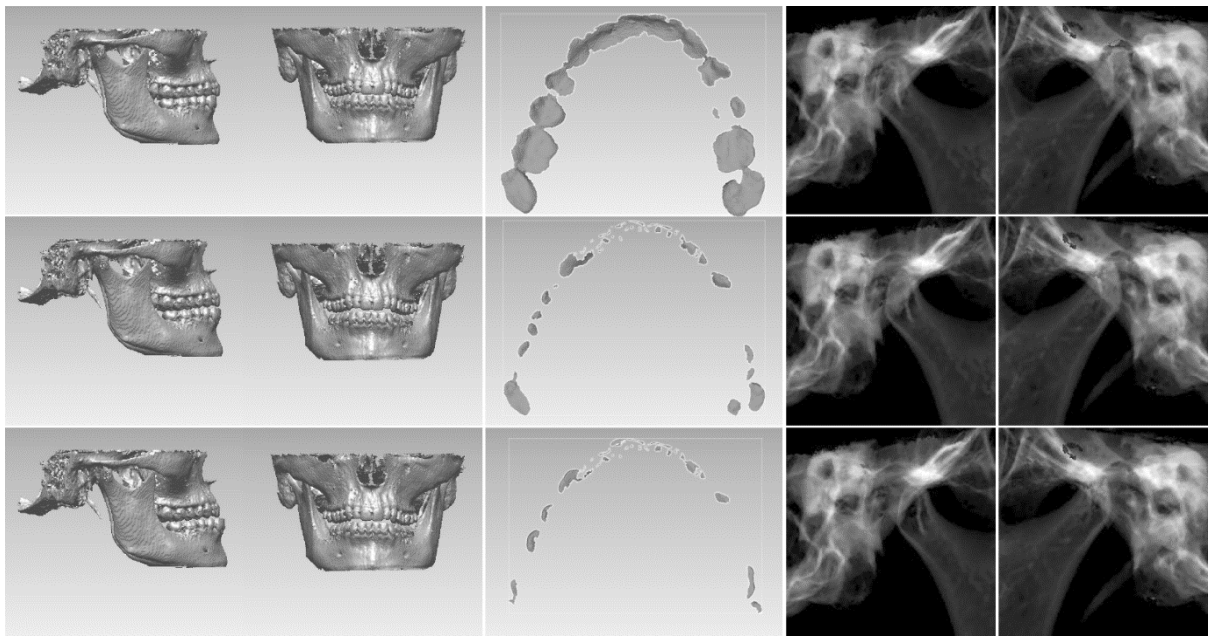

Fig. 13. Simulation of the mandible movement, the corresponding occlusion (see also [22]) and the virtual X-rays of the TMJ (see also [25]) in the static habitual occlusion (top row), incisal edge to edge position (middle row), and maximal protrusion (bottom row)

## 5. CONCLUSIONS

Simultaneous asessment of TMJ and occlusal area provides a powerful tool for the diagnosis of the discrepancy between the centric realtion and maximal intercuspation. The method of multimodal image integration is relatively new in medical applications and creates possibilities which were not available previously. However, to fully use its advantages, examination standards and movement strategies have to be developed. In order to define the individual movements, measurements have to be repeated, so statistics can be applied, and any irregularities related to pathological change detected.

All the medically relevant motions should be analyzed – such as the opening of the mouth, protrusion etc. Also, all positions need to be determined where tooth contact occurs – e.g. by analyzing the motions of biting and bruxing.

Simultaneous analysis of two key areas: occlusion and the TMJ working conditions opens new diagnostic possibilities. For each phase of motion, the tooth contact areas can be indicated. Non-physiological contacts can be identified, and their impact on the function of the TMJs studied. Location of contact points on individual teeth can be tracked and examined for any assymetries. The data obtained constitutes the premises for the creation ofq a patient-specific articulator, and can be exploited in the planning of occlusal splint therapy.

Additionally, perspectives arise of preparing patient specific motion models, which further can be used for planning appliances and therapy of bruxism.

A number of images are brought into register over a series of steps; the errors incurred in each of them propagate and accumulate. The errors depend on the precision and resolution of 3D scans. At the present moment, intraoral scans achieve a greater mesh density than face scans. In the process of automatic registration they impose a more precise alignment in the tooth area than on the face, because their contribution to the goal function is more significant. Another important source of error is the flexibility of soft tissues. Their changing shape makes automatic registration difficult; corresponding features have to be designated by hand (semiautomatically), in particular between the images with and without the occlusal bit and facial bow. The indirect registration method is also connected with errors as, because the long arm (distance between the facial bow and the teeth), a small registration error in one area can result in considerable error in the other.

Data analysis – there is still a need to more precisely state the meaning and the medical interpretation of the data. This work should be done in interdisciplinary research groups. For example, doctors expect movement to be expressed as a combination of translation and rotation, and they need to know which component dominates which phase of motion. They want to know where the rotation axis is situated in any given moment, and how it changes over time – problems which do not always have a unique and unambiguous solution. On the other hand, more complex data analysis can be difficult, if not impossible, to interpret medically.

## 6. PERSPECTIVES

With the rich arsenal of devices now available to obtain 2D, 3D and 4D images of biological objects and especially of the human body, combined with the algorithms which can integrate this imagery into a coherent record, avenues are open for the clinical, theoretical, and statistical analysis of such data. The long-established diagnostic standards based on ruler-and-protractor 2D measurements on flat X-ray images, such as Norma Frontalis or Norma Lateralis [16], should give way to richer and more comprehensive parametrizations of shape, geometry, and kinematics, less directly associated with the particular imaging devices, and more closely with the patient's condition. A major research challenge currently pursued in a number of research centres is the modelling and quantification of the behaviour of soft tissue during physiological movement, natural growth and aging, pathological processes, and medical intervention.

Future research should strive to further integrate the present models with information about the shape and movement of non-rigid tissues, including muscular activity. Statistical analysis of electromyographic [7][8] and electroencephalographic signals can complement the mechanical considerations with data representing the natural control mechanisms and effectuators. Importantly, multivariate cumulants of odd order can be used to measure an asymmetry of the signal. For an algorithm of efficient computation of such cumulants one can refer to [7].
This in turn should lead to better understanding and prevention of pathological processes such as bruxism or malocclusion, and their consequences, including tooth abrasion, fracture, gingival recession, tooth loss, muscular tension and pain.